\documentclass[letterpaper, 10 pt, conference]{ieeeconf}  % Comment this line out if you need a4paper

\IEEEoverridecommandlockouts                              % This command is only needed if
\usepackage{graphics} % for pdf, bitmapped graphics files
\usepackage{epsfig} % for postscript graphics files
\usepackage{mathptmx} % assumes new font selection scheme installed
\usepackage{times} % assumes new font selection scheme installed
\usepackage{amsmath}
\usepackage{amssymb}
\usepackage{caption}
\usepackage{cite}
\usepackage{color}
\usepackage{psfrag}
\usepackage{latexsym, amsmath, subfigure, color, amsfonts, amssymb,graphicx}
\usepackage{algorithm}
\usepackage{algorithmic}
\usepackage{cases}
\usepackage{acronym}
\usepackage{subfigure}
\usepackage{amsfonts}
\usepackage{setspace}

\title{\LARGE \bf
Robust Object Tracking with a Hierarchical Ensemble Framework
}

\author{Mengmeng Wang$^{1}$, Yong Liu$^{2}$ and Rong Xiong$^{2}$% <-this % stops a space
\thanks{*This work was supported in partby the National Natural Science Foundation Project of China under Project 61173123, in part by the Natural Science Foundation Project of ZhejiangProvince under Project LR13F030003, and in part by the Open ResearchProject of the State Key Laboratory of Industrial Control Technology, ZhejiangUniversity, China, under Project ICT1502.}% <-this % stops a space
\thanks{$^{1}$Mengmeng Wang is with the Institute of Cyber-Systems and Control, Zhejiang University, Zhejiang, 310027, China.}%
\thanks{$^{2}$Yong Liu and Rong Xiong are with the State Key Laboratory of Industrial Control Technology and Institute of Cyber-Systems and Control, Zhejiang University, Zhejiang, 310027, China (Yong Liu is the corresponding author of this paper, email: yongliu@iipc.zju.edu.cn).}%
}

\begin{document}

\maketitle
\thispagestyle{empty}
\pagestyle{empty}

%%%%%%%%%%%%%%%%%%%%%%%%%%%%%%%%%%%%%%%%%%%%%%%%%%%%%%%%%%%%%%%%%%%%%%%%%%%%%%%%
\begin{abstract}

Autonomous robots enjoy a wide popularity nowadays and have been applied in many applications, such as home security, entertainment, delivery, navigation and guidance. It is vital for robots to track objects accurately in real time in these applications, so it is necessary to focus on tracking algorithms to improve the robustness, speed and accuracy. In this paper, we propose a real-time robust object tracking algorithm based on a hierarchical ensemble framework which incorporates information including individual pixel features, local patches and holistic target models. The framework combines multiple ensemble models simultaneously instead of using a single ensemble model individually. A discriminative model which accounts for the matching degree of local patches is adopted via a bottom ensemble layer, and a generative model which exploits holistic templates is used to search for the object based on the middle ensemble layer as well as an adaptive Kalman filter. We test the proposed tracker on challenging benchmark image sequences. The experimental results demonstrate that the proposed tracker performs superiorly against several state-of-the-art algorithms, especially when the appearance changes dramatically and the occlusions occur.
\end{abstract}

%%%%%%%%%%%%%%%%%%%%%%%%%%%%%%%%%%%%%%%%%%%%%%%%%%%%%%%%%%%%%%%%%%%%%%%%%%%%%%%%
\section{INTRODUCTION}
Visual tracking is a well-studied problem in computer vision with a variety of applications such as surveillance, human motion analysis, robot guidance, human-computer interaction and so on. Recent attention has been focused to visual tracking in the robotic domains \cite{kolarow2012vision, klein2010adaptive}. However, due to the diverse environment and the complex motion of the robots, several tracking conditions such as occlusions, deformations, fast motion and background clutters remain difficult.
 \begin{figure}[!t]
\centering
\includegraphics[height=1.8in,width=3in,angle=0]{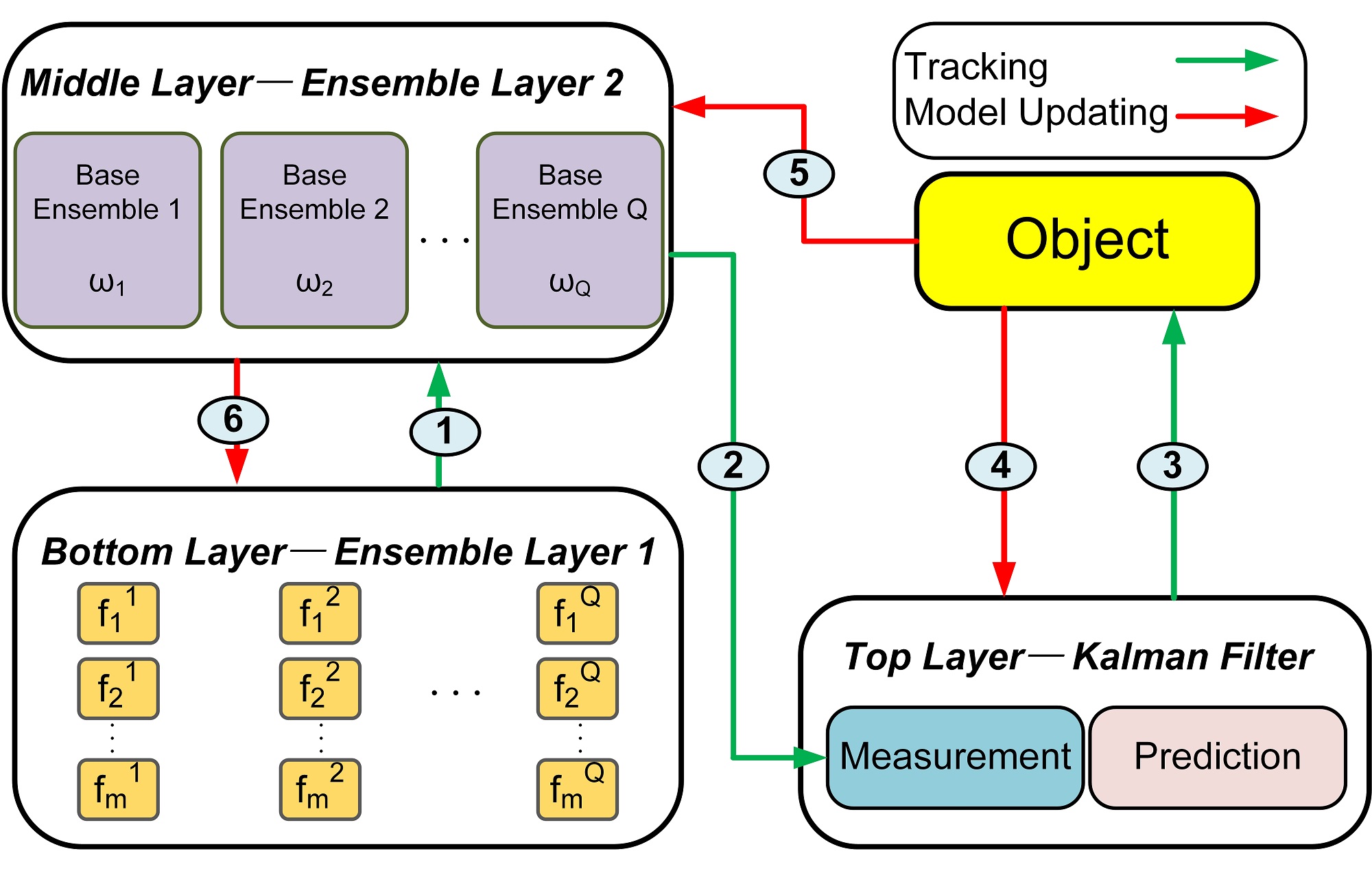}
\caption{An overview of the architecture of the layers. 1-combine the weak classifiers of each sub-patch to obtain the corresponding base ensembles and weights in the bottom layer; 2-combine the base ensembles to generate the measurement of the object in the middle layer; 3-employ an adaptive Kalman filter to increase the time consistency in the top layer; 4-update the top layer; 5-re-extract the sub-patches and update their weights in the middle layer; 6-update the parameters of weak classifiers in the bottom layer.}
\label{overview}
\end{figure}

There are three fundamental tracking components that are essential \cite{wu2015object} for improving performance of tracking: (1) the background information; (2) local appearance models; (3) motion models. This paper presents a hierarchical tracking framework which takes the above components into account. We model the object as an ensemble three-layer structure which can incorporate information including individual pixel features, the local patches and the target bounding box. The first component, i.e. the background information, is essential to overcome the background clutters due to the complexity of the environment. In our proposed method, we incorporate both the object and the background information into classifiers. For the second component, most of existing approaches \cite{li2015reliable, zhang2015structural}, which represent the target with a limited number of non-overlapping or regular local regions. So they may not cope well with the large deformations of the target. While our hierarchical tracker models the target with a series of overlapping and randomly sampled regions. We introduce the compressive sensing theory \cite{donoho2006compressed, candes2006stable} which significantly reduces the dimension of the pixel features in local regions. An overall schematic for the tracker is shown in Fig.\ref{overview}. For each sub-patch, we build a bottom ensemble layer which combines a collection of weak classifiers on the compressive features for the sub-patch into a strong classifier as a base ensemble. In the middle ensemble layer, we aggregate these base ensembles to generate the measurement of the target. As the robots move almost all the time when tracking an object, our approach needs to consider the third component and introduce an adaptive Kalman filter \cite{kalman1960new} in the top layer to consider the motion models and the temporal consistency in the target bounding box level. Above all, the contributions of our method are summarized as follows:
\begin{enumerate}
  \item We legitimately organize compressive features, overlapping sub-patches and holistic target models to capture the detailed appearance of the object;
  \item We propose a hierarchical ensemble framework that combines multiple ensemble models simultaneously instead of using a single ensemble model individually;
  \item We employ compressive sensing method to significantly reduce the feature dimensions so that our approach can handle colorful images without suffering from exponential memory explosion;
  \item We take the motion model into consideration to overcome the temporary occlusions, missing and false detections with an adaptive Kalman filter.
\end{enumerate}

In the experiment, we compare the proposed method against state-of-the-art tracking approaches which are feasible for robotic applications in terms of computational complexity and hardware requirements using an online object tracking benchmark \cite{wu2015object}. Our method obtains superior results compared with the state-of-the-art tracking approaches. The results also show that our method performs much better in the moving human tracking than other approaches for the conditions with occlusions, deformations, background clutters and scale variations.
\section{Related Work}
Recent tracking algorithms are developed in terms of three primary components: target representation, matching mechanism, and model update mechanism.

Target representation plays a pivotal role in visual tracking, and numerous representation schemes have been proposed. Several factors need to be considered for an effective appearance model in target representation. First, the features to represent the objects have many choices such as color histogram \cite{he2013visual}, superpixels \cite{xiao2015single}, Haar-like features \cite{hare2011struck, zhang2014fast,stalder2009beyond}, etc. Second, the templates to represent the objects can be global or local. Global templates \cite{zhang2014fast, kolarow2012vision} are easy to construct the object representation that contains information of the whole object. However, for the tracking problem of robots, holistic templates will have difficulty in handling significant appearance changes and deformations of the targets. While local templates \cite{liu2011robust, kalal2012tracking, li2015reliable} are more robust and flexible to these conditions. But the geometrical relationships for local patches remain tough since the environmental clutter, occlusions and partially similar objects can often distract such local patches and lead to drift.

Matching mechanism is used to classify candidate regions which are most similar to the target from background. There are two main streams of research on this: One is generative model which typically searches for the most similar candidate to the target within a neighborhood \cite{jia2012visual, kwon2010visual, henriques2012exploiting}. Another is discriminative model which poses the tracking problem as a binary classification task that determines the decision boundary for separating the target from the background \cite{stalder2009beyond, zhang2014fast, babenko2011robust, grabner2006real}.

Online model update mechanism is quite essential for robust visual tracking to deal with appearance variations. Addressing on this problem, Kalal et al. \cite{kalal2012tracking} develop a bootstrapping classifier to select positive and negative samples for model update. Grabner et al. \cite{grabner2008semi} formulate the update problem as a semi-supervised task where the classifier is updated with both labeled and unlabeled data. However, online boosting requires that the data should be independent and identically distributed. This is not always satisfied in visual tracking because the data are often temporally correlated.

In the proposed method, we adopt the compressive sensing theory to reduce the dimension of Haar-like features and this process is operated similarly to \cite{zhang2014fast}. We employ a joint representation which considers both global and local models of the target to better handle significant appearance changes, deformations, similar object identification and occlusions. Our local models are efficiently constructed with a number of overlapping and randomly sampled local patches and we re-extract the sub-patches at each time step to avoid the drifting caused by arbitrary sub-patch. We adopt a discriminative model via the bottom ensemble layer to account for the matching degree of local patches, and a generative model is used to seek for the object through the middle ensemble layer as well as an adaptive Kalman filter. For model update, we employ ensemble learning to update the patches and classifiers to capture appearance variations and reduce tracking drifts.
\section{Robust Object Tracking with a Hierarchical Ensemble Framework}

In this section, we give a detailed description of the proposed hierarchical ensemble tracking(HET) framework. It is composed of two ensemble layers and a Kalman filter layer. At each time step, we start with detecting several samples around each local patch and try to formulate the corresponding base ensemble for each sub-patch with several weak classifiers in the bottom ensemble layer. Second, we recover the target location in the middle ensemble layer by incorporating these base ensembles, and regard this location as the measurement to an adaptive Kalman filter. Third, we ascertain the ultimate object location at the current frame with a motion model and the measurement via the adaptive Kalman filter in the top layer. Finally, we update the model by re-extracting the local overlapping image sub-patches efficiently in the final target region with a random spatial layout and updating the parameters of weak classifiers for tracking in the next frame.

\subsection{Local Compressive Appearance Model}\label{Appearance Model}
The compressive sensing theory shows that if the dimension of the feature space is sufficiently high, these features can be projected to a randomly chosen low dimensional space which contains enough information to preserve most of the salient information of the original high-dimensional features through a random projection matrix  \cite{baraniuk2007compressive}. The signal can be recovered as long as the projection matrix $\bf{R}$ follows the Restricted Isometry Property (RIP)~\cite{candes2006stable}. Representing the object appearance by regions allows the proposed tracker to better handle occlusions and large appearance changes. The compressive appearance model also allows us to process a large number of regions in real-time.

In this paper, we build compressively sensed versions of sub-patches. Randomly extracted sub-patches are used and the relative location between sub-patches and the target bounding box are established when the tracking window is given by a detector or manual label at the first frame. Every sub-patch is represented by four components: a compressive feature vector ${{\bf{g}}^q}$, a classification score ${c_q}$, a relative location $\Delta {{\bf{p}}_q}$, where $\Delta {{\bf{p}}_q} = {\left[ {\Delta {x_q},\Delta {y_q}} \right]^{\rm{T}}}$ denotes the relative upper-left corner coordinate to upper-left corner of the target window, and the location of the sub-patch itself in the image space ${{\bf{p}}_q} = {\left[ {{x_q},{y_q}} \right]^{\rm{T}}}$. Denoted $q$-th sub-patch ${\lambda _q}$ as:
\begin{equation}\label{patches}
{\lambda _q} =  < {{\bf{g}}^q},{c_q},\Delta {{\bf{p}}_q},{{\bf{p}}_q} > .
\end{equation}

It is notable that the width and the height of each sub-patch are identical, denoted as $w$ and $h$, which are determined at beginning. After extracting these $Q$ local overlapping image sub-patches $\lambda  = \left\{ {{\lambda _1},{\lambda _2},...,{\lambda _Q}} \right\}$, where $Q$ denotes the number of sub-patches, for the $q$-th sub-patch, we sample $N$ sub-patches with the same size as the $q$-th sub-patch, whose Euclidean distances to the sub-patch is smaller than a threshold $\beta$ that is fixed through the sequence. These samples can form a matrix ${{\bf{S}}^q} = \left[ {{\bf{S}}_1^q,{\bf{S}}_2^q,...,{\bf{S}}_N^q} \right] \in {\mathbb{R} ^{w \times hN}}$. Then we present all samples as ${\bf{S}} = \left[ {{{\bf{S}}^1},{{\bf{S}}^2},...,{{\bf{S}}^Q}} \right] \in {\mathbb{R} ^{w \times hNQ}}$.

In order to find a kind of feature that is invariant to scale, we adopt a multiscale image representation that is often formed by convolving the input image with a Gaussian filter of different spatial variances and speed up the process via integral image method. We replace the Gaussian filter with rectangle filters for computation consideration \cite{zhang2014fast}. For $N$ samples of the $q$-th sub-patch ${{\bf{S}}^q}$, we obtain the feature matrix ${{\bf{H}}^q} = \left[ {{\bf{h}}_1^q,{\bf{h}}_2^q,...,{\bf{h}}_N^q} \right] \in {\mathbb{R} ^{n \times N}}$, the $k$-th column ${\bf{h}}_k^q \in {\mathbb{R} ^n}$, where $n \gg w \times h$ denotes the large multiscale feature vector of the $k$-th sample that is filtered with rectangle filters and concatenated as such a high-dimensional feature vector. Features of the total $N \times Q$ samples can denote as ${\bf{H}} = \left[ {{{\bf{H}}^1},{{\bf{H}}^2},...,{{\bf{H}}^Q}} \right] \in {\mathbb{R} ^{n \times NQ}}$.

We adopt a sparse random matrix ${\bf{R}} \in {\mathbb{R} ^{m \times n}}$, $m \ll n$ to reduce the original feature space $n$ into a lower-dimensional space $m$ such as ${{\bf{L}}^q} = \left[ {{\bf{l}}_1^q,{\bf{l}}_2^q,...,{\bf{l}}_N^q} \right] \in {\mathbb{R} ^{m \times N}}$ for $q$-th sub-patch.
Concatenating $Q$ local patches together, we obtain ${\bf{L}} = \left[ {{{\bf{L}}^1},{{\bf{L}}^2},...,{{\bf{L}}^Q}} \right] \in {\mathbb{R} ^{m \times NQ}}$, computed by
\begin{equation}\label{1}
{\bf{L}} = {\bf{RH}}
\end{equation}

\begin{figure}[!t]
\centering
\includegraphics[height=1.1in,width=2.5in,angle=0]{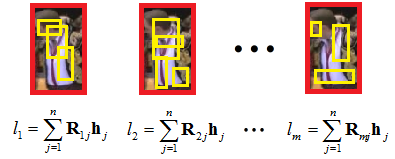}
\caption{An illustration for compressive representation for an arbitrary sample. Denote its high-dimensional feature vector as ${\bf{h}} \in {\mathbb{R} ^n}$. After the dimension reduction from $n$ to $m$, we get its $m$-dimensional feature vector ${\bf{l}} = {\left[ {{l_1},{l_2},...,{l_m}} \right]^{\rm{T}}} \in {\mathbb{R} ^m}$. Each element in ${\bf{l}}$ is linearly combined by the feature values of less than $Z$ rectangles(yellow) inside the sample region(red) and the coefficient of the combination is in the rows of ${\bf{R}}$. The feature values of each rectangle is actually the convolution from the corresponding rectangle filter that is the same size as the rectangle itself, i.e., the sum of gray values of all pixels inside it which can be computed very fast using the integral map.}
\label{compresive feature}
\end{figure}

A typical choice of such a measurement matrix is the random Gaussian matrix ${{\bf{R}}_{ij}} \sim {\rm N}\left( {0,1} \right)$. But when $n$ is huge, the computational loads are still heavy because the random Gaussian matrix is dense. Thus it is common to employ a very sparse random measurement matrix that satisfies a weaker property than RIP but almost as accurate as the conventional random Gaussian matrix~\cite{li2006very}, as (\ref{random metrix}), where ${{\bf{R}}_{ij}}$ denotes the element in the $i$-th row and $j$-th column of ${\bf{R}}$. This random matrix is fixed at the beginning and easy to compute for real-time tracking by fixing the maximum number $Z$ of nonzero elements to be a lower number. The scheme to produce the random matrix in this work is similar to \cite{zhang2014fast}. We illustrate the dimension reduction process in Fig.\ref{compresive feature}.
\begin{equation}\label{random metrix}
{{\bf{R}}_{ij}} = \left\{ \begin{array}{l}
\sqrt p ,\begin{array}{*{20}{c}}
{}&{with\;probability\;\frac{1}{{2p}}}
\end{array}\\
0,\begin{array}{*{20}{c}}
{ }&{with\;probability\;1 - \frac{1}{p}}
\end{array}\\
 - \sqrt p ,\begin{array}{*{20}{c}}
{}&{with\;probability\;\frac{1}{{2p}}}
\end{array}
\end{array} \right.
\end{equation}

\subsection{Classification via Ensemble Layers}

To link up the individual pixels with the local patches, we employ the naive Bayesian classifier to construct the pool of weak classifiers corresponding to each individual compressive feature in the bottom layer. We assume the compressive $m$-dimensional features of each sub-patch are independently distributed and build $m$ weak classifiers corresponding to these features by considering both the object and the background information. Since ${\bf{R}}$ is fixed during the tracking process, the way to compress the high dimensional features of samples stays consistent for all sub-patches. Let ${\bf{l}} = {\left[ {{l_1},{l_2},...,{l_m}} \right]^{\rm{T}}} \in {\mathbb{R} ^m}$ denote an arbitrary compressive sample, for the $i$-th compressive feature, the $i$-th classifier is constructed as follows:
\begin{equation}\label{classifier}
f\left( {{l_i}} \right) = \log \left( {\frac{{p\left( {y = 1|{l_i}} \right)}}{{p\left( {y = 0|{l_i}} \right)}}} \right) = \log \left( {\frac{{p\left( {{l_i}|y = 1} \right)p\left( {y = 1} \right)}}{{p\left( {{l_i}|y = 0} \right)p\left( {y = 0} \right)}}} \right),
\end{equation}
where $y \in \left\{ {0,1} \right\}$ is a binary variable which represents the sample label. We assume $p\left( {y = 1} \right)= p\left( {y = 0} \right)$ by sampling the same quantity of positive and negative samples at update step. The conditional distributions $p\left( {{l_i}|y} \right)$ are almost Gaussian due to the random projections of the high dimension features\cite{diaconis1984asymptotics}. Thus we have:
\begin{equation}\label{Gaussian}
p\left( {{l_i}|y = 1} \right)\; \sim {\rm N}\left( {\mu _i^1,\sigma _i^1} \right),p\left( {{l_i}|y = 0} \right)\; \sim {\rm N}\left( {\mu _i^0,\sigma _i^0} \right)\;,
\end{equation}
where $\mu _i^1\left( {\mu _i^0} \right),\sigma _i^1\left( {\sigma _i^0} \right)$ are the mean and standard deviation of the positive (negative) class.

\begin{figure}[!t]
\centering
\includegraphics[height=1.8in,width=3in,angle=0]{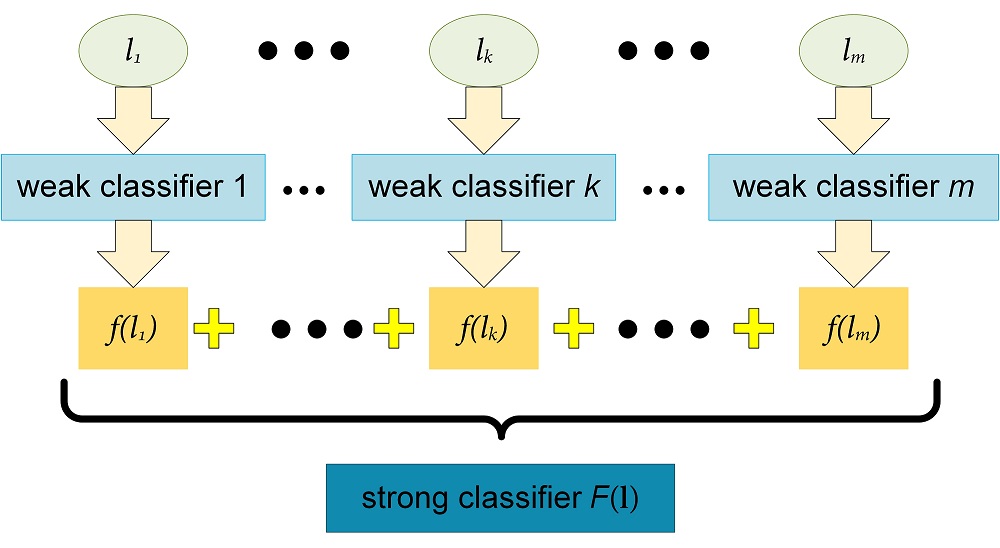}
\caption{The ensemble process in the bottom layer for an arbitrary sample ${\bf{l}} = {\left[ {{l_1},{l_2},...,{l_m}} \right]^{\rm{T}}} \in {\mathbb{R} ^m}$.}
\label{layer1}
\end{figure}

Then we introduce an ensemble strategy which combines the output of weak classifiers to create a strong classifier as a base ensemble to detect the sub-patches as shown in Fig.\ref{layer1}, denoted as
\begin{equation}\label{strongClassifier}
F\left( {\bf{l}} \right) = \sum\limits_{i = 1}^m {f\left( {{l_i}} \right)} ,
\end{equation}
For the $q$-th sub-patch, we seek its $N$ samples for matching and its matching score ${c_q}$ like:
\begin{equation}\label{classification}
\begin{array}{l}
{{\bf{g}}^q} = arg \mathop {\max }\limits_k \;\;F\left( {{\bf{l}}_k^q} \right),\;k = 1,...,N,\\
{c_q} = F\left( {{{\bf{g}}^q}} \right),
\end{array}
\end{equation}

We match all $Q$ sub-patches in the same way in the bottom layer and obtain the compressive feature of their optimal matching ${\bf{G}} = \left[ {{{\bf{g}}^1},{{\bf{g}}^2},...,{{\bf{g}}^Q}} \right] \in {\mathbb{R} ^{m \times Q}}$ and their scores ${\bf{c}} = {\left[ {{c_1},{c_2},...,{c_Q}} \right]^{\rm{T}}}$. In the ensemble learning field, it is often found that improved performance can be obtained by combining multiple models simultaneously like (\ref{strongClassifier}), instead of just using a single model individually \cite{rokach2010ensemble}.

\begin{figure}[!t]
\centering
\includegraphics[height=2.1in,width=3in,angle=0]{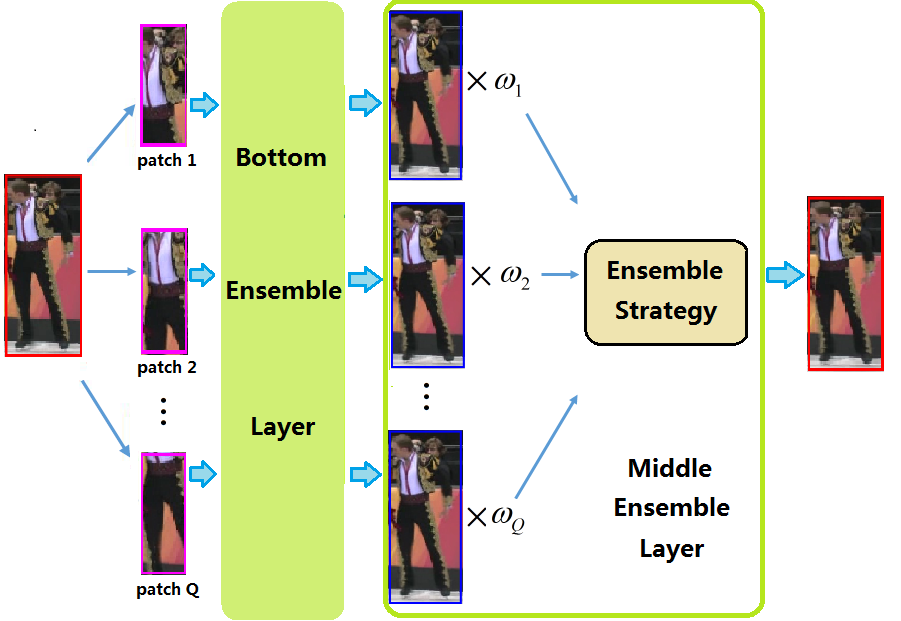}
\caption{The first column is the object at previous frame, second column denotes the randomly extracted sub-patches. Then transfer the sub-patches to the bottom ensemble layer to gain the base ensembles. The fourth column shows the scores and corresponding target candidates of the local patches which are the output of the base ensembles. Finally, we employ the proposed ensemble strategy to obtain the observation of object in the middle layer.}
\label{layer2}
\end{figure}

In the middle layer, we propose a novel ensemble strategy to acquire the observed location of the object from the base ensembles like Fig.\ref{layer2} via these $Q$ detected local patches.

Suppose the actual location of the object we are trying to predict is given by ${\rm{H}}\left( \lambda  \right)$, and ${y_i}\left( \lambda  \right) = \Delta {{\bf{p}}_i} + {{\bf{p}}_i}$ denotes the $i$-th hypothesis of object location obtained by the $i$-th detected sub-patch. The output of each sub-patch model can be written as the true value plus an error in this form:
\begin{equation}\label{error}
{y_i}\left( {\lambda} \right) = {\rm{H}}\left( {\lambda} \right) + {\varepsilon _i}\left( {\lambda} \right)
\end{equation}

To be convenient for comparison, we adapt the scores of sub-patches ${\bf{c}} = {\left[ {{c_1},{c_2},...,{c_Q}} \right]^{\rm{T}}}$ by the zero-mean normalization, then rescale them to ${\bf{\omega }} = {\left[ {{\omega _1},{\omega _2},...,{\omega _Q}} \right]^{\rm{T}}}$, ${\omega _i} \in \left[ {0.1,0.9} \right]$. ${\bf{\omega }}$ is regarded as the weights of candidates that obtained by the corresponding sub-patches. We update these weights adaptively for each new frame. The combined prediction is given by
\begin{equation}\label{weightAverage}
{y_{COM}} = \frac{1}{W}\sum\limits_{i = 1}^Q {{\omega _i}{y_i}\left( {\lambda} \right),\;\;W = {\omega _1} + ... + {\omega _Q}}
\end{equation}
The average sum-of-squares error then takes the form as follows:
\begin{equation}\label{sum-of-squares error}
{{\rm E}_{\lambda}}\left[ {{{\left( {{y_i}\left( {\lambda} \right) - {\rm{H}}\left( {\lambda} \right)} \right)}^2}} \right] = {{\rm E}_{\lambda}}\left[ {{\varepsilon _i}{{\left( {\lambda} \right)}^2}} \right]
\end{equation}
where ${{\rm E}_{\lambda}}\left[  \bullet  \right]$ denotes a frequentist expectation. The average error made by the sub-patch models acting individually is
\begin{equation}\label{average error}
{{\rm E}_{AV}} = \frac{1}{W}\sum\limits_{i = 1}^Q {{\omega _i}{{\rm E}_{\lambda}}\left[ {{\varepsilon _i}{{\left( {\lambda} \right)}^2}} \right]}
\end{equation}
We assume that the errors have zero mean and uncorrelated due to the sub-patches are randomly extracted. So we have:
\begin{equation}\label{assume}
{{\rm E}_{\lambda}}\left[ {{\varepsilon _i}\left( {\lambda} \right)} \right]{\rm{ = }}0,{{\rm E}_{\lambda}}\left[ {{\varepsilon _i}\left( {\lambda} \right){\varepsilon _j}\left( {\lambda} \right)} \right]{\rm{ = }}0,i \ne j
\end{equation}
The expected error from the combined prediction is computed by
\begin{equation}\label{expected error}
\begin{array}{l}
{{\rm E}_{COM}} = {{\rm E}_{\lambda}}\left[ {{{\left( {\frac{1}{W}\sum\limits_{i = 1}^Q {{\omega _i}{y_i}\left( {\lambda} \right)}  - {\rm{H}}\left( {\lambda} \right)} \right)}^2}} \right]\\
\;\;\;\;\;\;\;\; = {{\rm E}_{\lambda}}\left[ {\frac{1}{{{W^2}}}{{\left( {\sum\limits_{i = 1}^Q {{\omega _i}{\varepsilon _i}\left( {\lambda} \right)} } \right)}^2}} \right]\\
\;\;\;\;\;\;\;\; = \frac{1}{W}{{\rm E}_{\lambda}}\left[ {\frac{1}{W}\sum\limits_{i = 1}^Q {{\omega _i}{\omega _i}{\varepsilon _i}{{\left( {\lambda} \right)}^2}} } \right]\\
\;\;\;\;\;\;\; \le \frac{1}{W}{{\rm E}_{\lambda}}\left[ {\frac{1}{W}\sum\limits_{i = 1}^Q {{\omega _i}{\varepsilon _i}{{\left( {\lambda} \right)}^2}} } \right]\\
\;\;\;\;\;\;\; = \frac{1}{W}{{\rm E}_{AV}}
\end{array}
\end{equation}

We extract more than 10 sub-patches to ensure $W \ge 1$. The result suggests that the average error of a object model can be reduced weighted combining all the sub-patch models using (\ref{weightAverage}) on the key assumption (\ref{assume}) that the errors of each model are uncorrelated by randomly choose the sub-patches.

\subsection{Adaptive Kalman Filter}
The top layer builds an adaptive Kalman filter based on the two ensemble layers to estimate the optimal system state and target image velocity so that the proposed tracker can overcome the temporary occlusion, missing and false detections. We regard the observation result (\ref{weightAverage}) from the bottom and middle ensemble layers as the measurement. The discrete time system state and measurement at time $k$ are given by ${\bf{x}}\left( k \right) = {\left[ {x\left( k \right),y\left( k \right),{v_x}\left( k \right),{v_y}\left( k \right)} \right]^{\rm{T}}}$ and ${\bf{z}}\left( k \right) = {y_{COM}} = {\left[ {{x_o}\left( k \right),{y_o}\left( k \right)} \right]^{\rm{T}}}$, where $x\left( k \right),y\left( k \right),{x_o}\left( k \right),{y_o}\left( k \right)$ denote the center coordinate in the image space corresponding to system state and measurement at time $k$ respectively, ${v_x}\left( k \right),{v_y}\left( k \right)$ denote velocities in both two axis of system state. The state and measurement in the next time step $k$ + 1 is given by
\begin{equation}\label{Kalman(k+1)}
\begin{array}{l}
{\bf{x}}\left( {k + 1|k} \right) = {\bf{A}}\left( {k + 1|k} \right){\bf{x}}\left( {k|k} \right) + {\bf{\delta }}\left( {k + 1} \right),\\
{\bf{z}}\left( {k + 1} \right){\rm{ = }}{\bf{Bx}}\left( {k + 1|k} \right) + {\bf{\nu }}\left( {k{\rm{ + }}1} \right),\\
{\rm{ }}{\bf{A}}\left( {k + 1|k} \right) = \left[ {\begin{array}{*{20}{c}}
1&0&{\Delta t}&0\\
0&1&0&{\Delta t}\\
0&0&1&0\\
0&0&0&1
\end{array}} \right],{\bf{B}} = \left[ {\begin{array}{*{20}{c}}
1&0\\
0&1
\end{array}} \right],
\end{array}
\end{equation}
where ${\bf{A}}\left( {k + 1|k} \right)$ is modeled according to the Newton's equation of motion, ${\Delta t}$ is the time between two frames, ${\bf{\delta }}\left( {k + 1} \right) $ and ${\bf{\nu }}\left( {k{\rm{ + }}1} \right)$ are assumed to be white Gaussian noises with zero mean and covariance matrixes ${\bf{Q}}\left( {k} \right),{\bf{R}}\left( {k} \right)$ respectively. To achieve an adaptive Kalman filter, we take the mean of normalized scores in the middle layer to update these two covariance matrixes every frame like (\ref{Q_update}) and (\ref{R_update}). We ascertain the ultimate object location at the current frame in the top layer with this adaptive Kalman filter.
\begin{equation}\label{Q_update}
{\bf{Q}}\left( {k + 1} \right) = \left( {\frac{1}{Q}\sum\limits_{i = 1}^Q {\omega _i^{k + 1}} } \right){\bf{Q}}\left( 0 \right)
\end{equation}
\begin{equation}\label{R_update}
{\bf{R}}\left( {k + 1} \right) = \left( {1 - \frac{1}{Q}\sum\limits_{i = 1}^Q {\omega _i^{k + 1}} } \right){\bf{R}}\left( 0 \right)
\end{equation}

\subsection{Model Update}
It is important to update the target model continuously for robust tracking in the face of various difficult environment. The proposed method updates the hierarchical model via three mechanisms: re-extracting the sub-patches according to the object that we have found at the current frame, choosing the sub-patches that need to be updated and adjusting the parameters of the weak classifiers in the bottom layer. The update process is also shown in the Fig.\ref{overview}.

Once we find the object at the current frame, we need to correct the locations of all sub-patches in the middle layer due to the drift of the detection process. The way to re-extract the randomly overlapping sub-patches is fixed at the first frame. After that, we compress the features of these new sub-patches and put them into the weak classifiers in the bottom layer to obtain the updated scores ${{\bf{c}}^{new}} = {\left[ {c_1^{new},c_2^{new},...,c_Q^{new}} \right]^{\rm{T}}}$. We assume that $c_j^{new}$ is Gaussian, and the sub-patches to be updated are those whose scores satisfy
\begin{equation}\label{patchUpdate}
c_j^{new} \in \left( {{\mu _c} - {\sigma _c},{\mu _c} + {\sigma _c}} \right),j = 1,2,...,Q,
\end{equation}
where ${\mu _c},{\sigma _c}$ are mean and standard deviation of the scores.

Then, for the $j$-th chosen sub-patch, we extract $N$ positive samples whose Euclidean distances to the sub-patch are smaller than a threshold value $\beta$ and $N$ negative samples whose Euclidean distances to the sub-patch are bigger than a threshold value $\pi$ that is fixed at beginning. We update the parameters of its $i$-th weak classifier in (\ref{Gaussian}) like
\begin{equation}\label{parameterUpdate}
\begin{array}{l}
\mu _i^1 = \lambda \mu _i^1 + \left( {1 - \lambda } \right){\mu ^1}\\
\sigma _i^1 = \sqrt {\lambda {{\left( {\sigma _i^1} \right)}^2} + \left( {1 - \lambda } \right){{\left( {{\sigma ^1}} \right)}^2} + \lambda \left( {1 - \lambda } \right){{\left( {\mu _i^1 - {\mu ^1}} \right)}^2}}
\end{array}
\end{equation}
where ${\mu ^1}$, ${{\sigma ^1}}$ denote mean and standard deviation of the $N$ positive samples. And $\mu _k^0,\sigma _k^0$ are updated in a similar way.

\section{Experiments}
In this section, we show the experimental results of our method. Firstly, we present the implement details of the proposed tracker and the evaluation criteria to quantitatively assess the performance. Secondly, we validate the joint representation of our hierarchical ensemble framework with the base method. Thirdly, we compare our tracker to three most similar methods which are famous in the visual tracking field. Fourthly, we compare our method with 8 state-of-the-art methods which are feasible for robotic applications in terms of computational complexity and hardware requirements. Finally, we demonstrate that our tracker performs excellently for moving human tracking which is crucial for the tracking applications of robots.

\subsection{Implementation Details}

The proposed algorithm is implemented in Matlab(R2013a) and runs at 30 frames per second on an Intel i7-4790 machine with 3.6GHz CPU and 8GB RAM. For each sequence, the location of the target object is manually labeled at the first frame. For all reported experiments, we employ 150 weak classifiers in the bottom ensemble layer and randomly generate 11 sub-patches that are located inside the object and whose width and height are three quarters of the size of the object. We set learning rate $\lambda {\rm{ = }}0.85$,  maximum number of nonzero elements $Z=4$ in random matrix $\bf{R}$ and thresholds $\beta {\rm{ = }}20$, $\pi {\rm{ = }}2\beta$ in all experiments.

In the experiment, we employ two evaluation criteria to quantitatively assess the performance of the trackers including the average overlap rate and the center location error. Given the tracked bounding box $RO{I_T}$ and the ground truth bounding box $RO{I_{G}}$, we use the detection criterion in the PASCAL VOC challenge\cite{everingham2010pascal}, $score = \frac{{area\left( {RO{I_T} \cap RO{I_G}} \right)}}{{area\left( {RO{I_T} \cup RO{I_G}} \right)}}$ to evaluate the success rate.

\subsection{Comparison with the base method}
\begin{figure}[!t]
\centering
\subfigure [basketball]{\includegraphics[height=1.1in,width=1.6in,angle=0]{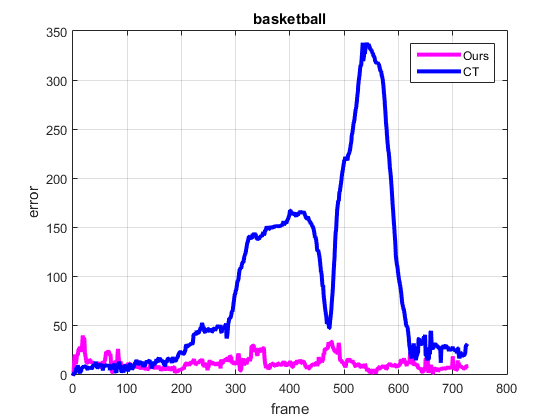}}
\subfigure [woman]{\includegraphics[height=1.1in,width=1.6in,angle=0]{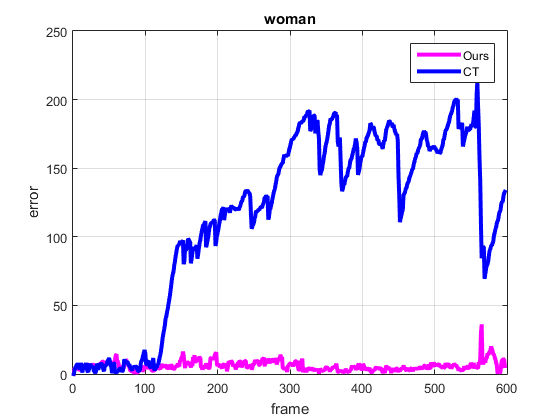}}
\subfigure [david3]{\includegraphics[height=1.1in,width=1.6in,angle=0]{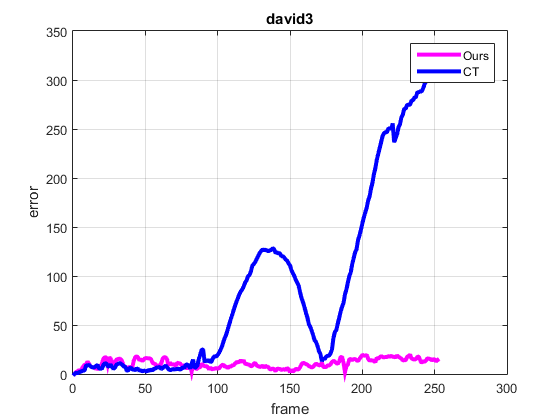}}
\subfigure [jogging-1]{\includegraphics[height=1.1in,width=1.6in,angle=0]{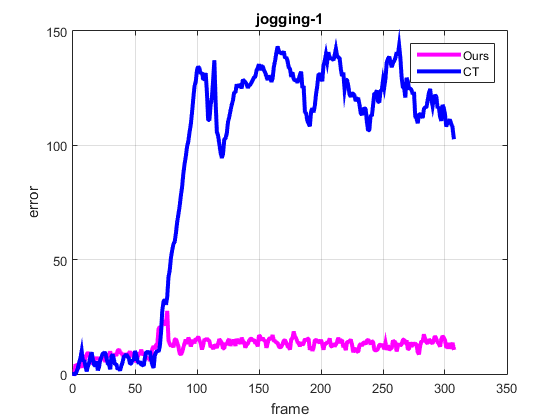}}
\caption{Pixel center location error of our method and the base method CT at each frame on four video sequences. Our method tracks the objects more accurately than CT on the four videos.}
\label{compareCT}
\end{figure}
Compressive tracking(CT)\cite{zhang2014fast} employs the compressive sensing theory to compress the appearance models. It is reasonable to consider CT as our base method since the way to compress a image sub-patch is almost the same. In the bottom layer of our method, we build compressively sensed versions of sub-patches, while CT presents objects by the compressive appearance models globally.

However, it's insufficient to present the holistic object by a single appearance model just like CT especially in the case of tracking non-rigid objects. So we adopt the joint representation which considers both global and local models of the targets to better handle significant appearance changes, deformations and occlusions. As shown in Fig.\ref{compareCT} and Fig.\ref{overall}, our method obtains more accurate tracking performances than the base method and it outperforms CT by 24\% for the success plots and by 38.3\% for the precision plots.

\subsection{Comparison with similar methods}

\begin{figure}[!t]
\centering
\subfigure [jogging-2]{\includegraphics[height=1.1in,width=1.6in,angle=0]{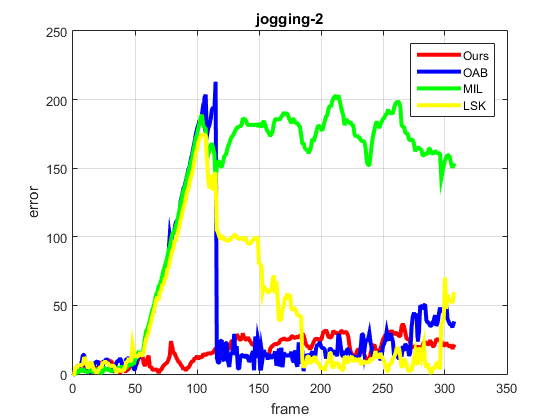}}
\subfigure [basketball]{\includegraphics[height=1.1in,width=1.6in,angle=0]{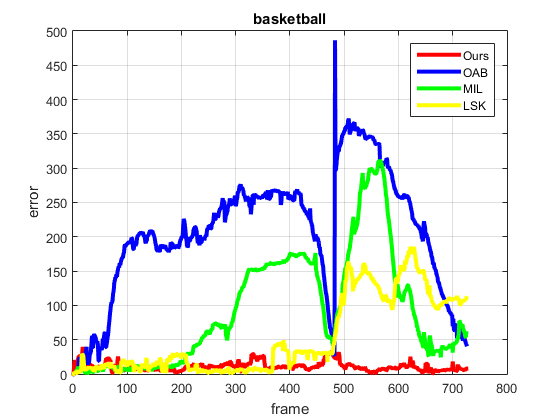}}
\subfigure [football]{\includegraphics[height=1.1in,width=1.6in,angle=0]{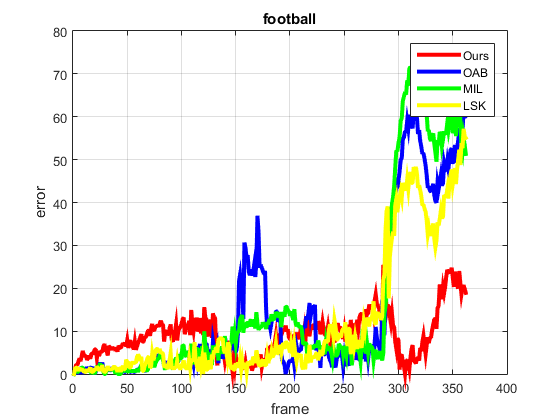}}
\subfigure [woman]{\includegraphics[height=1.1in,width=1.6in,angle=0]{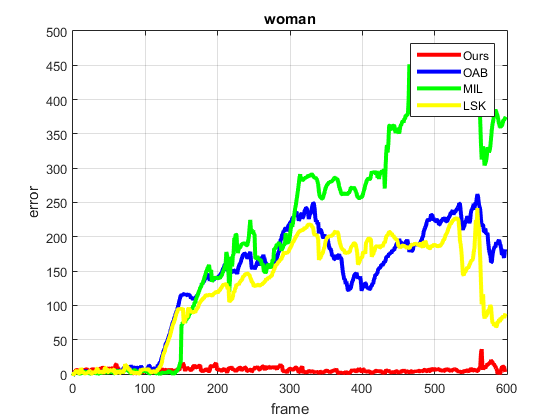}}
\caption{Pixel center location error of our method and the three similar methods at each frame on four video sequences. Our method tracks the objects more robustly than the three methods on these videos.}
\label{compare_sim}
\end{figure}

There are three methods LSK\cite{liu2011robust}, OAB\cite{grabner2006real} and MIL\cite{babenko2011robust} that are most similar to our tracker in recent years. The proposed method outperforms them as shown in Fig.\ref{compare_sim} and Fig.\ref{overall}.

\begin{figure*}[!t]
\centering
\includegraphics[height=3in,width=7in,angle=0]{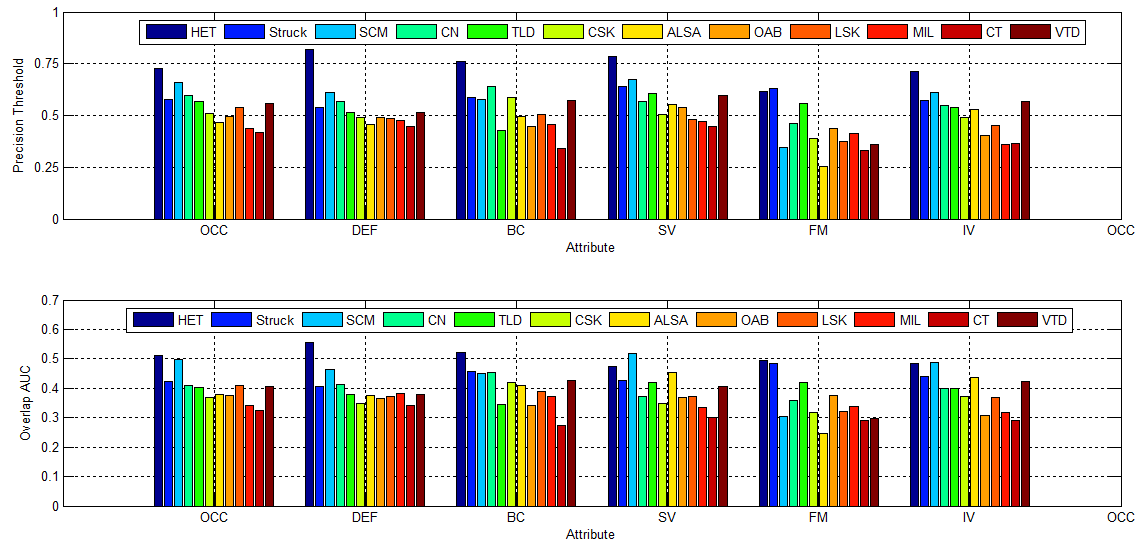}
\caption{Precision threshold and overlap AUC of 6 attributes on the 20 benchmark videos.}
\label{bar_attribution}
\end{figure*}

LSK proposes a robust tracking algorithm with a local sparse appearance model which combines a static sparse dictionary with a sparse coding histogram. This method outperforms several sparse representation methods according to \cite{wu2015object}. However, LSK neglects the temporal consistency in the target bounding box level while we take this into consideration by employing an adaptive Kalman filter. Therefore our method is more robust to occlusions than LSK, as shown in Fig.\ref{bar_attribution}.

OAB and MIL are both boosting-based algorithms similar with ours. Our ensemble technique is much easier than the boosting of the two methods. However, they characterize the objects by global templates while we adopt both local representations and holistic templates. Thus we can better handle the deformations and occlusions, as shown in Fig.\ref{bar_attribution}.

\subsection{Comparison with State-of-the-arts}

\begin{figure}[!t]
\centering
\subfigure [Precision]{\includegraphics[height=1.3in,width=1.6in,angle=0]{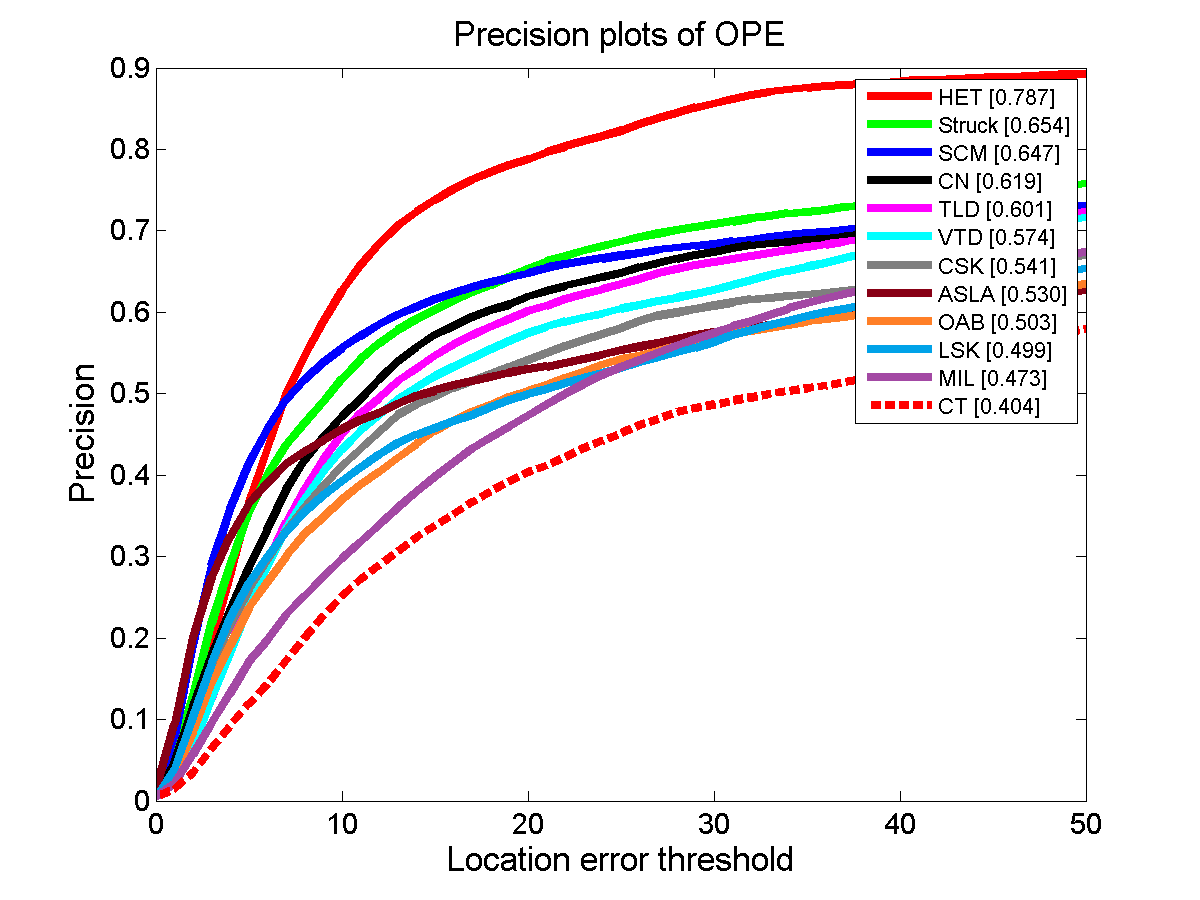}}
\subfigure [Success rate]{\includegraphics[height=1.3in,width=1.6in,angle=0]{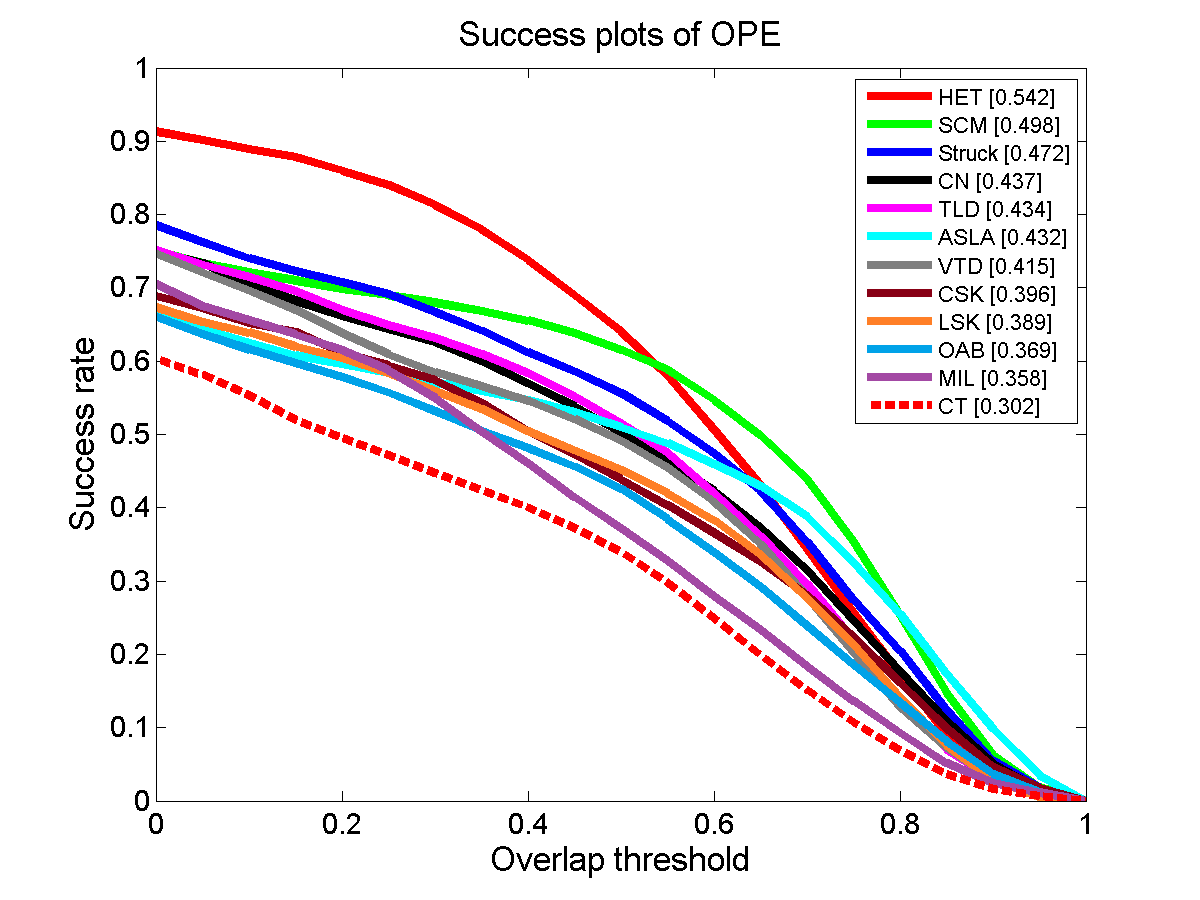}}
\caption{Precision and success rate plots of overall performance comparison for the 50 videos in the benchmark.}
\label{overall}
\end{figure}

For comparison, we run 11 state-of-the-art algorithms with the same initial positions of targets. These algorithms are CT\cite{zhang2014fast}, CN\cite{danelljan2014adaptive}, Struck\cite{hare2011struck}, TLD\cite{kalal2012tracking}, ASLA\cite{jia2012visual}, CSK\cite{henriques2012exploiting}, OAB\cite{grabner2006real},	MIL\cite{babenko2011robust}, LSK\cite{liu2011robust}, SCM\cite{zhong2014robust} and VTD\cite{kwon2010visual}. For fair evaluation, we evaluate the proposed HET against those methods using the source codes provided by the authors with adjusted parameters. We examine the effectiveness of the proposed approach on an online object tracking benchmark\cite{wu2015object} tested with 50 sequences that cover most challenging tracking scenarios such as illumination variations, scale variations, occlusions, deformations, etc.
\begin{figure*}[!t]
\centering
\subfigure {\includegraphics[height=0.7in,width=1in,angle=0]{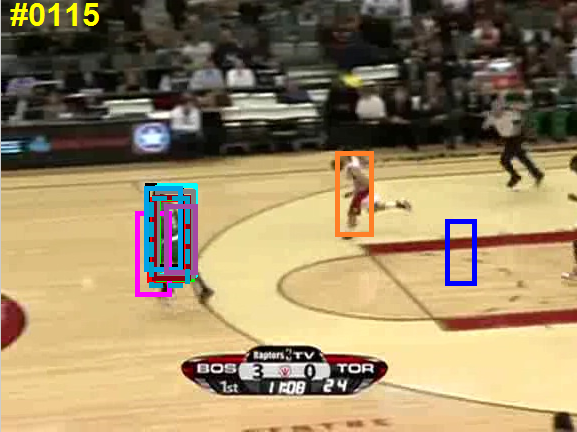}}
\subfigure {\includegraphics[height=0.7in,width=1in,angle=0]{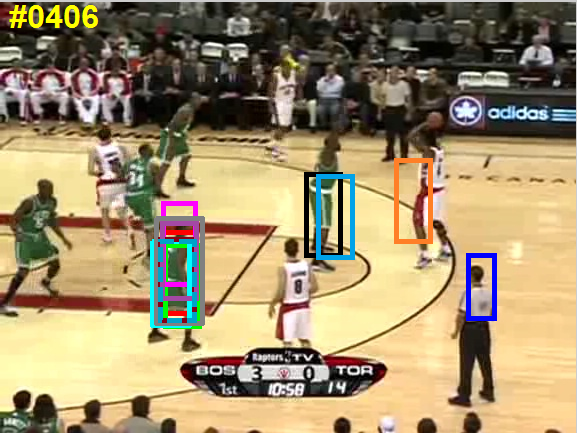}}
\subfigure {\includegraphics[height=0.7in,width=1in,angle=0]{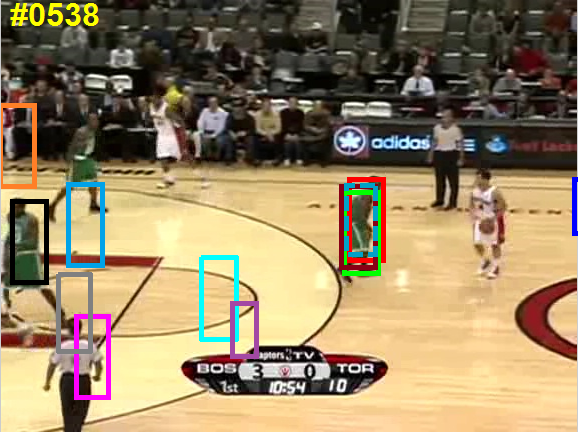}}
\subfigure {\includegraphics[height=0.7in,width=1in,angle=0]{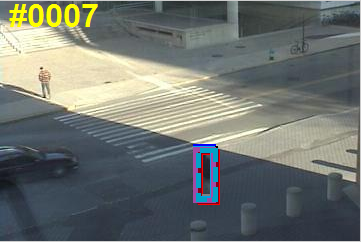}}
\subfigure {\includegraphics[height=0.7in,width=1in,angle=0]{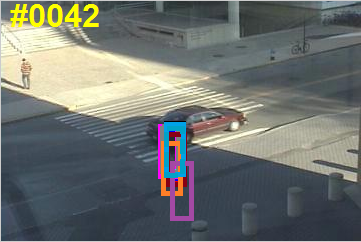}}
\subfigure {\includegraphics[height=0.7in,width=1in,angle=0]{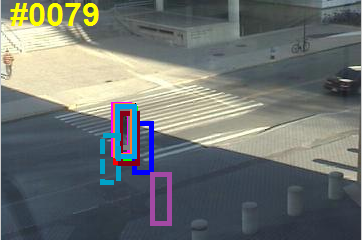}}
\subfigure {\includegraphics[height=0.7in,width=1in,angle=0]{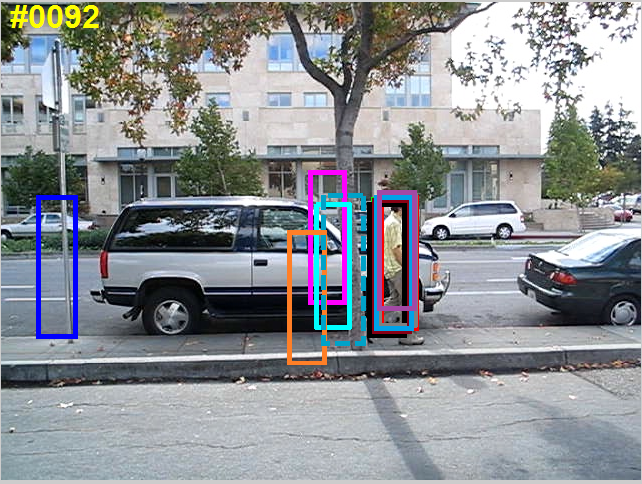}}
\subfigure {\includegraphics[height=0.7in,width=1in,angle=0]{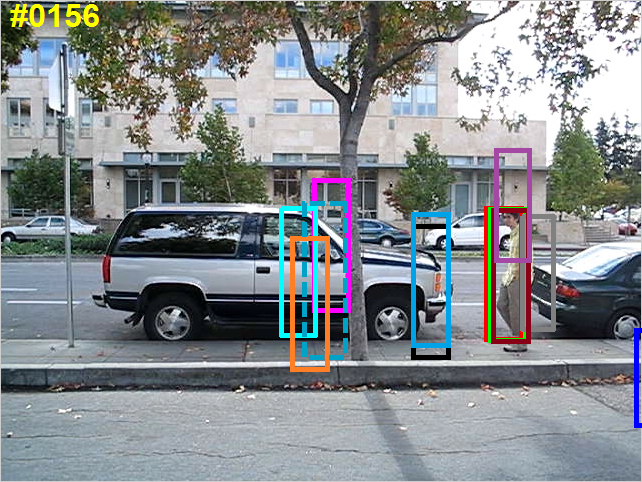}}
\subfigure {\includegraphics[height=0.7in,width=1in,angle=0]{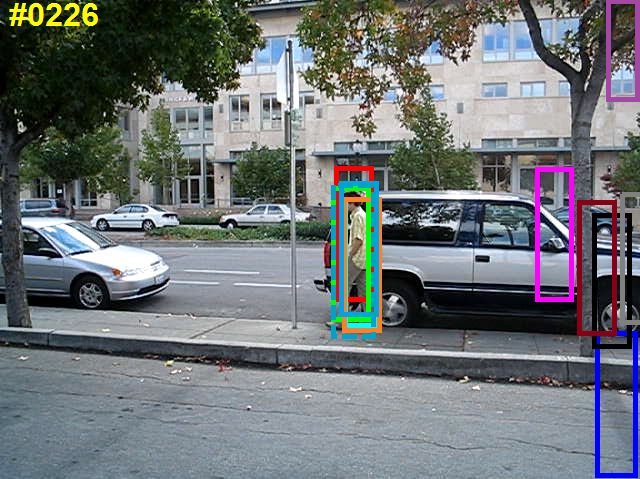}}
\subfigure {\includegraphics[height=0.7in,width=1in,angle=0]{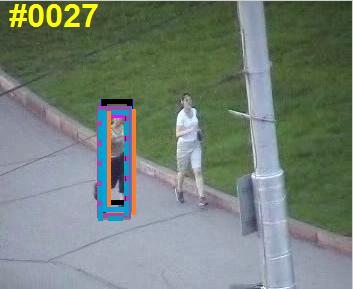}}
\subfigure {\includegraphics[height=0.7in,width=1in,angle=0]{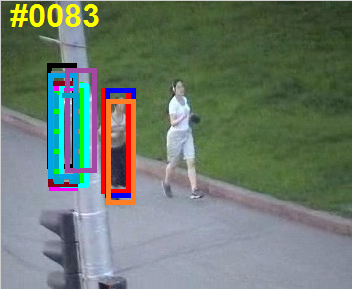}}
\subfigure {\includegraphics[height=0.7in,width=1in,angle=0]{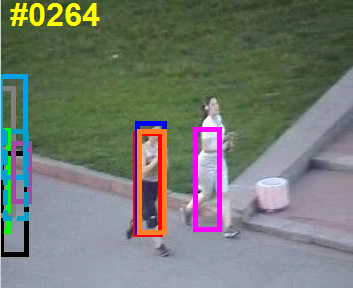}}
\subfigure {\includegraphics[height=0.7in,width=1in,angle=0]{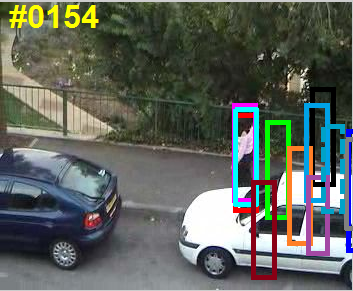}}
\subfigure {\includegraphics[height=0.7in,width=1in,angle=0]{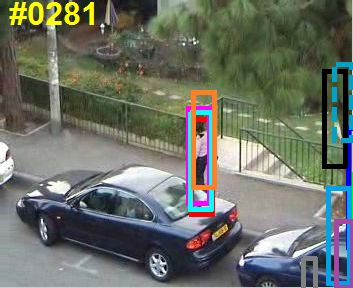}}
\subfigure {\includegraphics[height=0.7in,width=1in,angle=0]{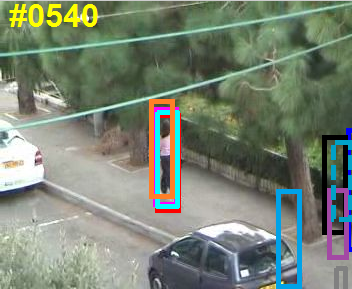}}
\subfigure {\includegraphics[height=0.7in,width=1in,angle=0]{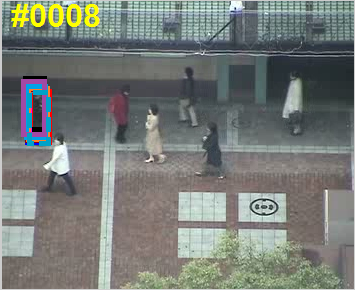}}
\subfigure {\includegraphics[height=0.7in,width=1in,angle=0]{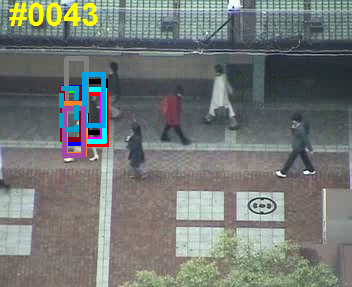}}
\subfigure {\includegraphics[height=0.7in,width=1in,angle=0]{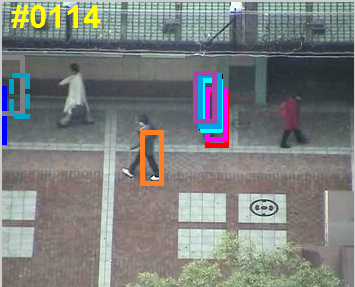}}
\subfigure {\includegraphics[height=0.18in,width=6in,angle=0]{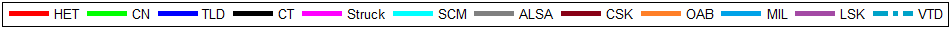}}
\caption{Tracking results of 12 trackers (denoted in different colors and lines) on 6 image sequences. Frame indexes are shown in the top
left of each figure in yellow color. Results are best viewed on high-resolution displays.}
\label{shots}
\end{figure*}

Note that we don't compare with some excellent methods such as MEEM\cite{zhang2014meem}, TGPR\cite{gao2014transfer}, MUSTer\cite{hong2015multi}, etc, because they are not fast enough in robotic applications. For instance, the average time of MUSTer cost on the benchmark\cite{wu2015object} is 0.287s/frame on a cluster node (3.4GHz, 8 cores, 32GB RAM). We also not consider some convolutional neural network based methods like FCNT\cite{wang2015visual} and HCF\cite{ma2015hierarchical} because they require powerful GPUs to run the algorithms while still in a low frame rate. Although the performances of these trackers may be a little better than ours, their computational costs and hardware requirements are impracticable for robots.

The comparison results of precision plots and success plots of OPE on benchmark are shown on Fig.\ref{overall}. As for the other top-ranking trackers in the benchmark, the results show that the proposed method achieves the best average performance. The performance of our approach can be attributed to the efficient ensemble methods on sub-patches with a spatial layout combining the adaptive Kalman filter.

\subsection{Human body tracking for robots}

In particular, we find that the proposed HET performs excellently for moving human tracking which is crucial for the tracking applications of robots. We choose 20 videos whose target objects are moving human bodies and includes the challenging conditions like occlusions, deformations, fast motions, background clutters, etc. Due to space limitations, we only show some shots of 6 videos among the 20 videos in Fig.\ref{shots}. We can see when the human body objects undergoing large deformations or occlusions, other methods almost lose the objects except the proposed HET.

Due to the suppleness structure of human limbs and the flexibility of human movements, the most challenging factors in moving body tracking are deformations and occlusions. It is obvious that splitting up a human object into local parts can make the appearance models more flexible. The local compressive appearance models of the proposed HET actually do these things. We build compressively sensed versions of local patches in the bottom layers which allows the proposed tracker to better handle occlusion and large appearance change. The tracking results on these moving human videos with occlusions(OCC), deformation(DEF), background clutter(BC), scale variations(SV), fast motion(FM) and illumination variation(IV) attributes based on the precision and success rate metrics are persuasive, as shown in Fig.\ref{bar_attribution}. Our method almost ranks the first on these attributes according to the two criteria.

The robustness and realtime performance to the human body tracking makes HET suitable for many robotic applications such as human-computer interaction, home service robots, robot teaching systems and unmanned vehicles.

\section{Conclusion}
In this paper, we propose a novel hierarchical ensemble framework, where the representations of the target candidates are localized and compressed. We incorporate information including individual pixel features, local patches and holistic target models. The multiple ensemble layers exploit the intrinsic relationship not only between the individual pixel features and local patches, but also between the patches and the target candidates. In the bottom layer, the base ensembles are created using linear combinations of outputs from the base weak classifiers. A diverse collection of base ensembles are systematically combined in order to generate a more strong ensemble classifier in the middle layer and the scores of the local patches are normalized to produce a vector of weights of the base ensembles. Experimental results with evaluations against several state-of-the-art methods on challenging image sequences demonstrate the robustness of the proposed HET tracking algorithm. Since our method is real-time, general and robust, we plan to apply it to the tracking tasks of robots. In particular, the proposed HET is very efficient for moving human tracking, we can apply this to many applications such as unmanned vehicles, robot teaching system, etc.

\bibliographystyle{ieeetr}
\bibliography{cite}

\end{document}